\documentclass{article}

\usepackage{microtype}
\usepackage{graphicx}
\usepackage{subfigure}
\usepackage{booktabs} 

\usepackage{hyperref}

\usepackage{caption}

\usepackage[accepted]{icml2023}

\usepackage{amsmath}
\usepackage{amssymb}
\usepackage{mathtools}
\usepackage{amsthm}

\usepackage[capitalize,noabbrev]{cleveref}

\theoremstyle{plain}

\theoremstyle{definition}

\theoremstyle{remark}

\usepackage[textsize=tiny]{todonotes}

\usepackage{amsmath}
\usepackage{lipsum}

\usepackage{setspace}
\makeatletter
\renewcommand\subsubsection{\@startsection{subsubsection}{3}{\z@}{-0.04in}{0.0001in}  {\normalsize\bf\raggedright}}
\renewcommand\paragraph{\@startsection{paragraph}{4}{\z@}{0.001ex plus 0.001ex minus .001ex}{-1em}{\normalsize\bf}}
\makeatother

\usepackage{multirow}
\usepackage{tablefootnote}
\usepackage{pifont}
\definecolor{Gray}{gray}{0.86}
\definecolor{LightGray}{gray}{0.94}
\usepackage{xcolor}
\definecolor{darkred}{HTML}{bb0000}
\definecolor{darkpink}{HTML}{ca00ff}

\usepackage{amsmath,amsfonts,bm}

\def\eqref#1{equation~\ref{#1}}

\def\1{\bm{1}}

\DeclareMathAlphabet{\mathsfit}{\encodingdefault}{\sfdefault}{m}{sl}
\SetMathAlphabet{\mathsfit}{bold}{\encodingdefault}{\sfdefault}{bx}{n}

\usepackage{mathtools}
\usepackage{xspace}

\usepackage{inconsolata}

\newcommand{\methodname}{RA-CM3\xspace}

\icmltitlerunning{Retrieval-Augmented Multimodal Language Modeling}

\begin{document}
\setlength{\abovedisplayskip}{6pt}
\setlength{\belowdisplayskip}{6pt}

\twocolumn[
\icmltitle{Retrieval-Augmented Multimodal Language Modeling}



\icmlsetsymbol{equal}{*}
\icmlsetsymbol{thanks}{*}

\begin{icmlauthorlist}
\icmlauthor{Michihiro Yasunaga,\!}{su,thanks}
\icmlauthor{Armen Aghajanyan,\!}{fb}
\icmlauthor{Weijia Shi,\!}{uw}
\icmlauthor{Rich James,\!}{fb}
\icmlauthor{Jure Leskovec,\!}{su}
\icmlauthor{Percy Liang}{su}
\icmlauthor{Mike Lewis,\!}{fb}
\icmlauthor{Luke Zettlemoyer,\!}{uw,fb}
\icmlauthor{Wen-tau Yih}{fb}
\end{icmlauthorlist}

\icmlaffiliation{su}{Stanford University}
\icmlaffiliation{fb}{Meta AI}
\icmlaffiliation{uw}{University of Washington}

\icmlcorrespondingauthor{Michihiro Yasunaga}{myasu@cs.stanford.edu}

\icmlkeywords{Machine Learning, ICML}
\vskip 0.2in

{
\hspace{-2mm}\includegraphics[width=1.03\textwidth]{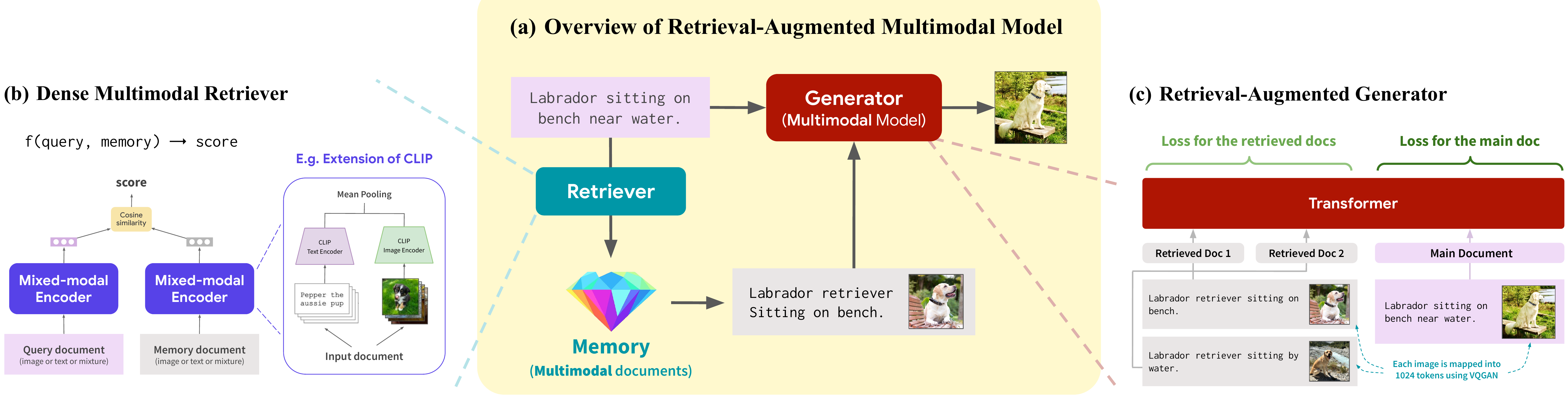}
    \vspace{-6mm}
    \captionof{figure}{\textbf{Our approach, retrieval-augmented multimodal modeling}.
    \textbf{(a)} Overview: given an input multimodal document, we use a \textbf{retriever} to retrieve relevant multimodal documents from an external memory, and use the \textbf{generator} to refer to the retrieved documents and make predictions for the input (e.g., generate the continuation).
    \textbf{(b)} The multimodal retriever is a dense retriever with a mixed-modal encoder that can encode mixture of text and images (\S \ref{sec:approach-retrieval}).
    \textbf{(c)} The retrieval-augmented generator uses the CM3 Transformer architecture, and we prepend the retrieved documents to the main input document that we feed into the model
    (\S \ref{sec:approach-generator}).
    }
    \label{fig:overview} }

\vskip 0.2in
]



\printAffiliationsAndNotice{}  

\begin{abstract} 
Recent multimodal models such as DALL-E and CM3 have achieved remarkable progress in text-to-image and image-to-text generation. 
However, these models store all their knowledge (e.g., the appearance of the Eiffel Tower) in the model parameters, requiring increasingly larger models and training data to capture more knowledge.
To integrate knowledge in a more scalable and modular way, we propose a retrieval-augmented multimodal model, which enables a base multimodal model (\textit{generator}) to refer to relevant text and images fetched by a \textit{retriever} from external memory (e.g., documents on the web).
Specifically, for the retriever, we use a pretrained CLIP, and for the generator, we train a CM3 Transformer on the LAION dataset.
Our resulting model, named Retrieval-Augmented CM3 (\methodname), is the first multimodal model that can retrieve and generate both text and images.
We show that \methodname significantly outperforms baseline multimodal models such as DALL-E and CM3 on both image and caption generation tasks (12 FID and 17 CIDEr improvements on MS-COCO), while requiring much less compute for training ($<\!30\%$ of DALL-E).
Moreover, we show that \methodname exhibits novel capabilities, such as faithful image generation and multimodal in-context learning (e.g., image generation from demonstrations).
\end{abstract}

\vspace{-6mm}
\begin{table*}
\begin{center}
\scalebox{0.78}{
\begin{tabular}{lccccc}
\toprule
\textbf{Approach} & \textbf{Model type} & \textbf{Image generation} & \textbf{Text generation} & \textbf{Retrieval} & \textbf{In-context learning}\\
\midrule
DALL-E, Parti \scalebox{0.65}[0.67]{(\citeauthor{ramesh2021zero,yu2022scaling})} & Autoregressive & \ding{52} &  & \\
DALL-E 2, Imagen \scalebox{0.65}[0.67]{(\citeauthor{ramesh2022hierarchical,saharia2022photorealistic})} & Diffusion & \ding{52} & \\
Re-Imagen \scalebox{0.65}[0.67]{(\citeauthor{reimagen})} & Diffusion & \ding{52} &  &  \ding{52}  \\
Flamingo, MetaLM \scalebox{0.65}[0.67]{(\citeauthor{alayrac2022flamingo, hao2022language})} & Autoregressive & & \ding{52} & & \ding{52} \\
MuRAG \scalebox{0.65}[0.67]{(\citeauthor{chen2022murag})} & Autoregressive & & ~\,\ding{52}$^\dagger$ & \ding{52} &  \\
CM3 \scalebox{0.65}[0.67]{(\citeauthor{aghajanyan2022cm3})}\ & Autoregressive & \ding{52} & \ding{52} &  & \\
\midrule
\textbf{\methodname (Ours)} & Autoregressive & \ding{52} & \ding{52} & \ding{52} & \ding{52}  \\
\bottomrule
\end{tabular}}
\vspace{-1mm}
\caption{\textbf{Comparison with other multimodal models}. Our \methodname is the first retrieval-augmented model that can perform both image and text generation. \methodname also exhibits strong in-context learning abilities thanks to the proposed retrieval-augmented training (\S \ref{sec:approach-generator}). $^\dagger$Focus on question answering.
\label{tbl:method_comparison}
}
\end{center}\vspace{-3mm}
\end{table*}

\section{Introduction}
Recent multimodal models have achieved remarkable progress in image and text generation. 
DALL-E \cite{ramesh2021zero} and Parti \cite{yu2022scaling} perform image generation from text, Flamingo \cite{alayrac2022flamingo} performs text generation from images, and CM3 \cite{aghajanyan2022cm3} offers a unified Transformer model that generates both text and images.
Typically, these models store all their knowledge (e.g., the appearance of the Eiffel Tower) implicitly in the parameters of the underlying neural network, requiring a lot of parameters (e.g., 10--80B) and training data (e.g., 1--10B images) to cover all the knowledge.
This motivates the development of multimodal models that can refer to an external memory of knowledge (e.g., web data) for increased knowledge capacity. 
Access to external memory is useful to accommodate the growth and update of knowledge through time, and is especially helpful for tasks that involve entity knowledge, such as generating images for entity-rich captions like ``George Washington standing in front of the Eiffel Tower''. Reference to external memory may also offer benefits such as explainable and faithful predictions \cite{metzler2021rethinking}.

Recently, retrieval-augmented language models have shown promise in natural language processing (NLP) \cite{karpukhin2020dense,guu2020realm,lewis2020retrieval,borgeaud2022improving}.
Given input text, such a model uses a \textit{retriever} that retrieves relevant documents from an external memory, and uses a \textit{generator} to generate predictions given the retrieved documents.
However, these retrieval-augmented methods are studied originally for text, and extending them to the multimodal setting remains an open problem with challenges.
Specifically, we need to design a retriever and a generator that handle multimodal documents, consisting of \textit{both} images and text.
Several concurrent works study retrieval for multimodal data \cite{chen2022murag, reimagen}, but their generators are each limited to a single modality, either text generation or image generation (Table \ref{tbl:method_comparison}).

In this work, we address the above challenge and present the first retrieval-augmented multimodal model that can retrieve and generate \textit{both} text and images.
As in Figure \ref{fig:overview}, our input data and external memory comprise a set of \textit{multimodal documents}, each of which is an arbitrary sequence of text/images (e.g., text, image, or their combinations like caption-image pair).
First, to obtain a {multimodal retriever}, we use the Dense Retrieval method \cite{karpukhin2020dense} with a mixed-modal encoder that can encode combinations of text and images (e.g., pretrained CLIP; \citealt{radford2021learning}). Given this retriever, we design a technique to retrieve diverse and informative documents for the input document.
Second, we design the {retrieval-augmented generator} based on the CM3 architecture \cite{aghajanyan2022cm3}, which is a Transformer sequence model capable of both text and image generation. Concretely, we prepend the retrieved documents as in-context examples to the main input document, and train the generator by optimizing token prediction loss jointly for the main document and retrieved documents.

We train our retrieval-augmented CM3 (\textit{\methodname}), using 150M text-image pairs from the LAION dataset \cite{schuhmann2021laion}. 
\methodname achieves strong performance on MS-COCO image and caption generation, significantly outperforming the baseline CM3 with no retrieval (12 FID and 17 CIDEr
improvements). It also outperforms existing models such as DALL-E and Flamingo, despite using fewer parameters ($<\!30\%$) and compute for training ($<\!30\%$).

We further demonstrate novel capabilities of \methodname (\S \ref{sec:qualitative-results}). 
First, it can perform faithful generation for tasks that require entity knowledge, for which existing models struggle (Figure \ref{fig:knowledge_intensive}, \ref{fig:rare_composition}).
Second, \methodname exhibits a multimodal in-context learning ability: it can perform controlled image generation by prompting with demonstration examples in context (Figure \ref{fig:control_gen}), and it can also perform few-shot image classification.
\methodname is the first model that can perform in-context learning for both text and image generation (Table \ref{tbl:method_comparison}).

More broadly, our work offers a general and modular retrieval augmentation framework for multimodal models, and opens up various research avenues, such as further advancement of multimodal retrievers and generators.

\section{Related work} 
We discuss related works in detail in \S \ref{sec:related_work}.

\section{Approach}
\label{sec:approach}

We present a retrieval-augmented multimodal model that can retrieve and generate both text and images.
As illustrated in Figure \ref{fig:overview}, given an input multimodal document (i.e., arbitrary sequence of text/images), we use a \textbf{retriever} that retrieves relevant multimodal documents from an external memory, and uses a \textbf{generator} to refer to the retrieved documents and make predictions for the input document (i.e., generate the continuation).
We design the multimodal retriever as a dense retriever with a mixed-modal encoder that can encode combinations of text and images (e.g., pretrained CLIP; \S \ref{sec:approach-retrieval}).
We build the retrieval-augmented generator using the CM3 Transformer architecture, and we prepend the retrieved documents to the main input document that we feed into the generator (\S \ref{sec:approach-generator}).
We describe how we train this model and use it for text-to-image or image-to-text generation in \S \ref{sec:approach-train}.
Notably, our resulting model, \textit{{Retrieval-Augmented CM3} (\textit{RA-CM3})}, is the first multimodal model that can retrieve and generate \textit{combinations} of text and images, which is the most general capability among existing multimodal models (Table \ref{tbl:method_comparison}).
Moreover, while we build on existing techniques such as CLIP and CM3, we are the first to establish a method to unify them into a performant retrieval-augmented model through extensive analyses of design choices (\S \ref{sec:exp-method-design}).

\subsection{Preliminaries}
\paragraph{Retrieval augmented language model.} The framework consists of a retrieval module $R$ and a generator module $G$ (e.g., language model).
The retrieval module $R$ takes an input sequence $x$ and an external memory of documents $\mathcal{M}$, and returns a list of documents $M\subseteq \mathcal{M}$.
The generator $G$ then takes the input sequence $x$ and the retrieved documents $M=(m_1, ..., m_K)$, and returns the target $y$, where $y$ is the continuation of $x$ in a typical language modeling task.

\paragraph{Causal masked multimodal model (CM3).}
CM3 \cite{aghajanyan2022cm3} is a Transformer decoder \cite{vaswani2017attention} model for multimodal documents. A \textit{multimodal document} is defined as an arbitrary sequence of text/images (e.g., text, image, or their combinations like caption-image pair). In particular, CM3 formats each multimodal document as an HTML sequence, such as ``\texttt{<img alt=[text] src=[image]>}'', where \texttt{[text]} is a sequence of text tokens, and \texttt{[image]} is a sequence of image tokens obtained by an image tokenizer such as VQGAN \cite{esser2021taming}, which maps a raw image into 1024 tokens.

At training time, CM3 either takes the original sequence as input (e.g., $x_\text{input}=$ ``Photo of a cat: \texttt{[image]}'') or converts it into an infilling instance by masking some spans and moving them to the end (e.g., $x_\text{input}=$ ``Photo of \texttt{<mask>}: \texttt{[image]} \texttt{<infill>} a cat''), and then optimizes the standard next token prediction loss for the input, $-\log p(x_\text{input})$.
This provides a flexible model that learns to perform infilling besides standard autoregressive generation.
In particular, the model can perform both image and text generation: for caption-to-image, CM3 generates a continuation from the prompt ``Photo of a cat:''. For image-to-caption, CM3 generates from the prompt ``Photo of \texttt{<mask>}: \texttt{[image]} \texttt{<infill>}''.

\paragraph{Our setup.}
We aim to generalize the retrieval-augmented language model framework to the multimodal setting.
Our input $x$ + target $y$ will be a multimodal document, and our memory $\mathcal{M}$ will be a set of multimodal documents.
We design the retrieval module $R$ for multimodal data (\S \ref{sec:approach-retrieval}), and design the multimodal generator $G$ based on CM3 (\S \ref{sec:approach-generator}).

\subsection{Multimodal retrieval}
\label{sec:approach-retrieval}

\paragraph{Dense retriever.}
A retriever $r$ takes a query $q$ (e.g., the input sequence $x$) and a candidate document $m$ from the memory $\mathcal{M}$, and returns a relevance score $r(q, m)$.
We follow the Dense Retrieval method \cite{karpukhin2020dense}, in which the retriever $r$ is a bi-encoder architecture,
\begin{align}
    r(q, m) = E_Q(q)^\top E_M(m)
\end{align}
where the query encoder $E_Q$ and memory encoder $E_M$ produce dense vectors for the query and memory document, respectively (Figure \ref{fig:overview}b). As our input and memory are multimodal documents, we let $E_Q$ and $E_M$ be \textbf{mixed-modal encoders} that encode a combination of text and images. While any mixed-modal encoders could be used in our framework, we find that a simple extension of CLIP \cite{ramesh2021zero} works well empirically, so we adopt it in our final system.
Concretely, as shown in Figure \ref{fig:overview}b (right), given a multimodal document, we split it into a text part and an image part, encode the two parts separately using off-the-shelf frozen CLIP text and image encoders, and then average the two, with the L2 norm scaled to 1, as the vector representation of the document. We use this encoding method for both $E_Q$ and $E_M$. Intrinsic evaluation of this CLIP-based retriever can be found in \S \ref{sec:retrieval_eval}.

Given this retriever $r$, we perform Maximum Inner Product Search (MIPS; \S \ref{sec:exp-train_setup}) over the memory to obtain a list of candidate documents sorted by the relevance score. We then sample the final $K$ retrieved documents from this list.

\paragraph{Retrieval strategy.}
\label{sec:approach-retrieval-strategy}
We discuss three key factors in obtaining/sampling informative retrieved documents for the generator in practice. 

\textit{\textbf{Relevance}}: The retrieved documents need to be relevant to the input sequence; otherwise, the retrieved documents do not provide useful information for modeling the main input sequence (see \S \ref{sec:exp-method-design} for the ablation study). The dense retriever score based on CLIP captures this relevance factor.

\textit{\textbf{Modality}}: While existing works on retrieval \cite{reimagen} typically retrieve either an image or text only for the generator, we find that retrieving a multimodal document that consists of both images and text leads to better generator performance (see \S \ref{sec:exp-method-design}). Our intuition is that a multimodal document can be more informative because the text and image within it can contextualize each other. Hence, in our final system, we retrieve the raw multimodal documents that keep both images and text for the generator.


\textit{\textbf{Diversity}}: 
We find that ensuring diversity in retrieved documents is important.
First, simply sampling or taking the top $K$ from the document list based on the relevance score can result in duplicate or highly similar images or text, leading to poor generator performance. This is especially important in the multimodal setting because even when two multimodal documents are not duplicates by themselves, the images or text contained in them can be duplicates, hurting the generator performance. To avoid redundancy, when we take documents from the top of the list, we skip a candidate if it is too similar (e.g., relevance score $> 0.9$) to the query or to the documents we already retrieved.
Second, to further encourage diversity, we also propose Query Dropout, which drops some tokens of the query used in retrieval (e.g., 20\% of tokens). This technique serves as regularization for training, and leads to further improvement in generator performance. Hence, our final system uses these two techniques (Avoid Redundancy + Query Dropout) for training, and uses Avoid Redundancy for inference.
See \S \ref{sec:exp-method-design} for detailed analysis.

\subsection{Multimodal generator}
\label{sec:approach-generator}
We use CM3 as the base of our multimodal generator $G$.
To incorporate the retrieved documents $M=(m_1, ..., m_K)$ into the generator, we prepend them to the main input sequence $x$, and feed the resulting sequence $(m_1, ..., m_K, x)$ to the Transformer (Figure \ref{fig:overview}c). In other words, the retrieved documents are in-context examples for the main input.

To train the generator, we optimize the following loss:
\begin{align}
    L &= L_{\text{main}} + \alpha L_{\text{retr}}
    \label{eq:loss}\\
    & = -\log p(x \!\mid\! m_1, ..., m_K) - \alpha \log p(m_1, ..., m_K)
\end{align}
where $L_{\text{main}}$ and $L_{\text{retr}}$ are the CM3 token prediction loss for the main input sequence $x$ and for the retrieved documents $(m_1, ..., m_K)$, respectively.
Here we propose optimizing the two loss terms jointly, with $\alpha \geq 0$.
Existing retrieval-augmented models (e.g., \citealt{lewis2020retrieval}) typically only optimize the loss for the main sequence, $L_{\text{main}}$ (i.e., $\alpha = 0$). However, as the Transformer computes logits for tokens in the retrieved documents when it computes logits for tokens in the main sequence, we can easily include the loss for the retrieved documents, $L_{\text{retr}}$. 
Thus, $\alpha>0$ offers an effect analogous to increasing the batch size (the number of tokens involved in optimization) without much extra compute, and boosts training efficiency. 
This technique is especially useful in the multimodal modeling setting, because each image takes many tokens (e.g., 1024 tokens), and $\alpha=0$ would throw away computation used for the image tokens in retrieved documents.
In practice, we find $\alpha=0.1$ works well. See \S \ref{sec:exp-method-design} for detailed analysis.

\subsection{Training and inference}
\label{sec:approach-train}
\paragraph{Training.}
Given a full input document $x$, we use either its text part or its image part as the query $q$ for retrieving documents (\S \ref{sec:approach-retrieval}). We then optimize the generator token prediction loss over the whole concatenated sequence (Equation \ref{eq:loss}) by standard teacher forcing. We only use the text or image part as the query because (1) retrieving documents based on the full input document could make the generator's token prediction task too easy during training,
and (2) this training setting is close to the typical inference scenarios of text-to-image and image-to-text generation.

Since our off-the-shelf CLIP-based retriever already performs well, we fix the retriever and only train the generator in this work.
An interesting future research direction would be the exploration of co-training or fine-tuning the retriever.

\paragraph{Inference.}
Our method takes an input sequence (prompt) $x$, uses $x$ as the query for retrieval, and then lets the generator take the retrieved documents as part of the input to decode the continuation of $x$.
For instance, for text-to-image generation, prompt $x$ takes the source caption, and the continuation will be the target image. For image-to-text, prompt $x$ takes the source image, and the continuation will be the target caption.
Thus, the retriever only uses the prompt as a query and never sees the ground-truth continuation $y$ to be evaluated, ensuring no information leakage.
\section{Experiments}
\label{sec:experiments}
To experiment with our proposed approach, we train models using the LAION mutlimodal dataset (\S \ref{sec:exp-train_setup}), and evaluate on the MS-COCO image and caption generation tasks (\S \ref{sec:exp-eval_setup}).
We show that our retrieval-augmented model (\methodname) significantly improves both image and text  generation performance (\S \ref{sec:exp-main_result}). 
We then analyze the scaling laws and key design choices of our model (\S \ref{sec:exp-scaling_law}, \ref{sec:exp-method-design}).
Finally, \S \ref{sec:qualitative-results} presents qualitative results and capabilities of our model, such as knowledge intensive generation and in-context learning.

\subsection{Training setup} 
\label{sec:exp-train_setup}
\paragraph{Data.}
To train our model, we use LAION \cite{schuhmann2021laion}, an open-sourced dataset that consists of text-image pairs collected from the web.
Following the preprocessing step of Stable Diffusion \cite{rombach2022high}, we cleaned a subset of LAION\footnote{We filter out images with watermark probability above 0.5, unsafe probability above 0.5, or resolution below 256 \!$\times$\! 256.} and obtained 150M text-image pairs in total. Following CM3, we format each text-image pair as an HTML document, ``\texttt{<img alt=[text] src=[image]>}'', where \texttt{[image]} is a sequence of 1024 image tokens obtained by tokenizing the raw image using VQGAN \cite{esser2021taming, gafni2022make}.
These 150M documents are used as our model's final training data. 

We also use the same 150M documents for our external memory $\mathcal{M}$.

\paragraph{Implementation.}
In our retrieval module $R$, we use the off-the-shelf CLIP model (\texttt{ViT-L/14}) \cite{radford2021learning} for both the query and memory encoders $E_Q$ and $E_M$.
We use FAISS \cite{johnson2019billion} to index the external memory $\mathcal{M}$ (Flat Index) and perform MIPS-based retrieval.

For our generator $G$, we use a Transformer \cite{vaswani2017attention} of 2.7B parameters. The sequence length is 4096, which can take up to 3 documents. For each input document $x$, we retrieve $K \sim \operatorname{Uniform}(\{0,1,2\})$ documents and prepend them to $x$. At inference time, we may also retrieve and add $K\!>\!2$ documents via ensemble (see \S \ref{sec:result-oneshot}).\\[0.8mm]
The model is trained from scratch for five days on 256 A100 GPUs.
Our implementation is in PyTorch \citep{pytorch} using Metaseq \citep{zhang2022opt}. We use model parallelism over 4 GPUs and a batch size of 16 sequences per GPU. The optimization uses a linear learning rate decay with 1500 warmup steps, a peak learning rate of 1e-4, a gradient clipping of 1.0, and the Adam optimizer with $\beta_1=0.9$, $\beta_2=0.98$ \citep{kingma2015adam}.

\begin{table}[t]
\begin{center}
\setlength\tabcolsep{2pt}
\resizebox{0.49\textwidth}{!}{
\begin{tabular}{llcccccc@{}}
\toprule
&\multicolumn{1}{l}{\multirow{2}{*}{\textbf{Approach}}} && \multicolumn{1}{c}{\multirow{2}{*}{\textbf{Model type}}} && \multicolumn{2}{c}{\textbf{MS-COCO FID ($\downarrow$)}}\\
\cmidrule{6-7}
&& && & Not finetuned & Finetuned \\
\midrule
&Retrieval Baseline  && -   &&  17.97 & -\\
\midrule
&KNN-Diffusion~\scalebox{0.8}{\cite{ashual2022knn}}\!  && Diffusion & & 16.66 & - &    \\
&Stable Diffusion~\scalebox{0.8}{\cite{rombach2022high}}\!\!\!  && Diffusion & & 12.63 & - &    \\
&GLIDE~\scalebox{0.8}{\cite{nichol2021glide}}  && Diffusion & & 12.24 & - &    \\
&DALL-E 2~\scalebox{0.8}{\cite{ramesh2022hierarchical}}  && Diffusion & & 10.39 & - &   \\
\midrule
&Imagen~\scalebox{0.8}{\cite{saharia2022photorealistic}}  && Diffusion & & {7.27} & - &  \\
&Re-Imagen~\scalebox{0.8}{\cite{reimagen}}  && Diffusion & & {6.88} & 5.25 &  \\
\midrule
&DALL-E (12B)~\scalebox{0.8}{\cite{ramesh2021zero}}  && \scalebox{0.95}[1]{Autoregressive}\!\!  & & $\sim$28 & - &    \\
&CogView (4B)~\scalebox{0.8}{\cite{ding2021cogview}}  && \scalebox{0.95}[1]{Autoregressive}\!\! & & 27.1 & - &  \\
&CogView2 (6B)~\scalebox{0.8}{\cite{ding2022cogview2}}  && \scalebox{0.95}[1]{Autoregressive}\!\! & & 24.0 & 17.7 &  \\
&Parti (20B)~\scalebox{0.8}{\cite{yu2022scaling}}  &&  \scalebox{0.95}[1]{Autoregressive}\!\!  && 7.23 & 3.22 \\
\midrule
&Vanilla CM3 &&  \scalebox{0.95}[1]{Autoregressive}\!\!  && 29.5 & - \\
&\textbf{\methodname (2.7B) (Ours)}  &&  \scalebox{0.95}[1]{Autoregressive}\!\!  && \textbf{15.7} & - \\
\bottomrule
\end{tabular}
}\vspace{-1mm}
\caption{\textbf{Caption-to-image generation performance on MS-COCO}.
Our retrieval-augmented CM3 significantly outperforms the baseline CM3 with no retrieval, as well as other models such as DALL-E (12B parameters). Moreover, our model achieves strong performance with much less training compute than existing models; see Figure \ref{fig:compute} for details.
}
\label{tbl:coco_image_results}
\end{center}\vspace{-2mm}
\end{table}
\begin{figure}[!t]
    \centering
    \includegraphics[width=0.35\textwidth]{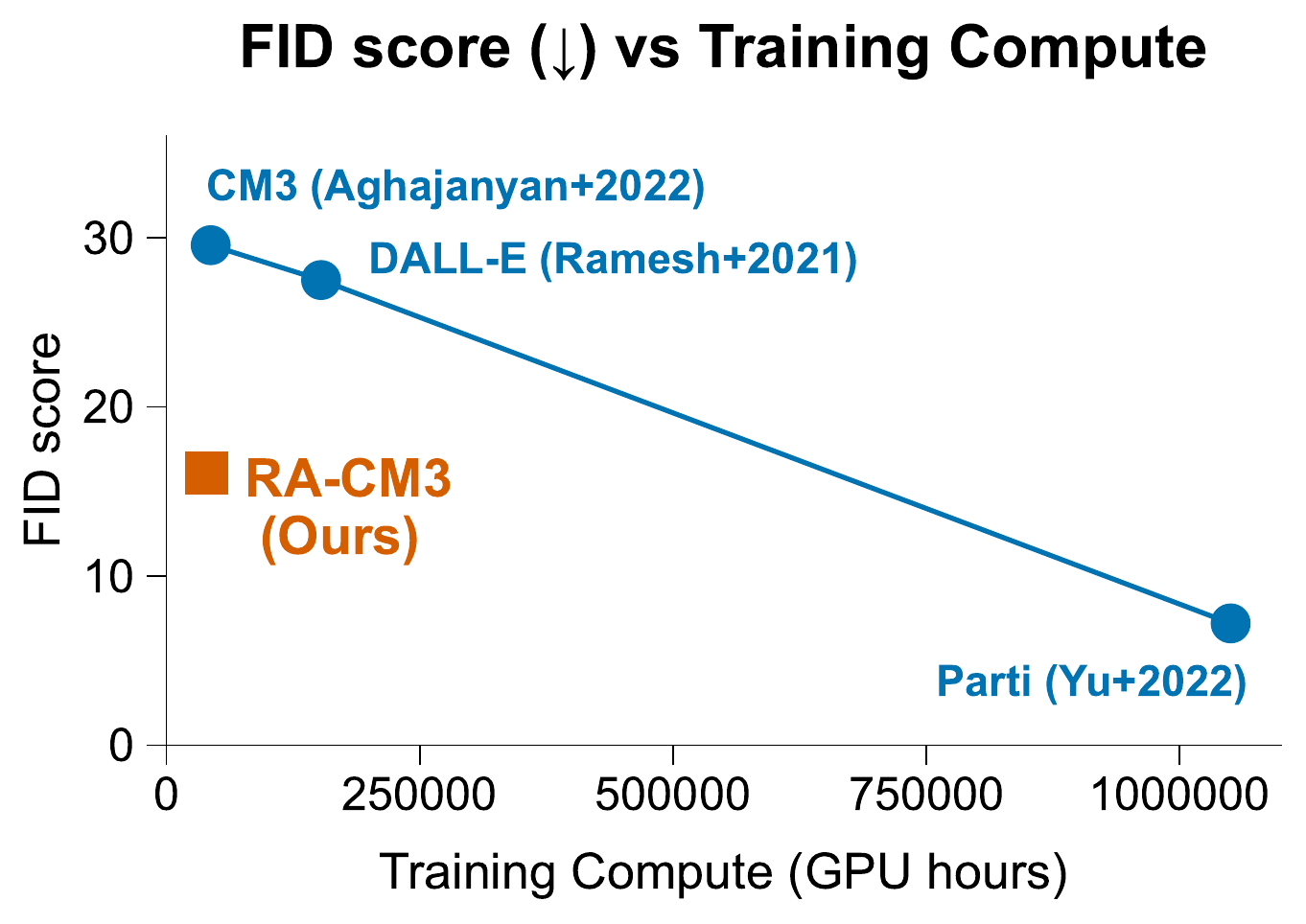}
    \vspace{-2mm}
    \caption{\textbf{Image generation quality vs training compute} for our \methodname model and baseline models. $x$-axis is the amount of training compute used in terms of A100 GPU hours. $y$-axis is the MS-COCO FID score (the lower, the better).
    Our retrieval-augmented method achieves significantly better training efficiency than existing works under a similar autoregressive Transformer paradigm (e.g., CM3, DALL-E, Parti). 
    }
    \label{fig:compute}
\end{figure}

\paragraph{Baseline.} 
For our baseline, we train a vanilla CM3 with no retrieval augmentation, using the same model architecture, training data, and amount of compute, for {fair comparison}. Since \methodname's external memory consists of the same training data, the total information accessible to \methodname and vanilla CM3 are controlled to be the same.

\subsection{Evaluation setup}
\label{sec:exp-eval_setup}
For the main evaluation, we use the standard benchmark, MS-COCO  \cite{lin2014microsoft}, to evaluate both text-to-image and image-to-text generation. 
We evaluate our trained model with no further finetuning.

For text-to-image, following prior works \cite{ramesh2021zero, nichol2021glide}, we generate images for the MS-COCO validation set captions and measure the FID score \cite{heusel2017fid} against ground-truth images. To generate an image for each caption, we sample 10 images from the model and then take the top image based on the CLIP score \cite{radford2021learning} with respect to the input caption, as done in \citet{aghajanyan2022cm3}.

For image-to-text, following prior works \cite{alayrac2022flamingo}, we generate captions for the MS-COCO validation set images and measure the CIDEr score \cite{vedantam2015cider} against ground-truth captions.
To generate an caption for each image, we sample 32 captions from the model and take the top caption based on perplexity \citep{fried2022incoder}.

\begin{table}
\begin{center}
\scalebox{0.75}{
\begin{tabular}{lc}
\toprule
\textbf{Approach} & \textbf{CIDEr ($\uparrow$)} \\
\midrule
Retrieval Baseline & 84.1 \\
\midrule
DALL-E\textsuperscript{Small} \scalebox{0.8}{\cite{lucidrains-dalle}}   & 20.2  \\
ruDALL-E-XL \scalebox{0.8}{\cite{ru-dalle}}  & 38.7   \\
minDALL-E \scalebox{0.8}{\cite{kakaobrain2021minDALL-E}}       & 48.0  \\
X-LXMERT \scalebox{0.8}{\cite{cho2020x}}      & 55.8  \\
Parti \scalebox{0.8}{\cite{yu2022scaling}} & {83.9} \\
Flamingo (3B; 4-shot) \scalebox{0.8}{\cite{alayrac2022flamingo}} & {85} \\
Flamingo (80B; 4-shot) \scalebox{0.8}{\cite{alayrac2022flamingo}} & {103} \\
\midrule
Vanilla CM3 &  71.9  \\
\textbf{\methodname (2.7B) (Ours)} &  \textbf{89.1}  \\
\bottomrule
\end{tabular}}
\vspace{-1mm}
\caption{\textbf{Image-to-caption generation performance on MS-COCO} (with no finetuning).
Our retrieval-augmented CM3 significantly outperforms the baseline CM3 with no retrieval. Moreover, our model outperforms other strong models such as Parti (20B parameters) and Flamingo (3B; 4-shot), despite using just $\sim$3B parameters and 2-shot in-context examples.
\label{tbl:coco_caption_results}
}
\end{center}\vspace{-2mm}
\end{table}

\begin{figure*}[!t]
    \hspace{-2mm}
    \includegraphics[width=1.03\textwidth]{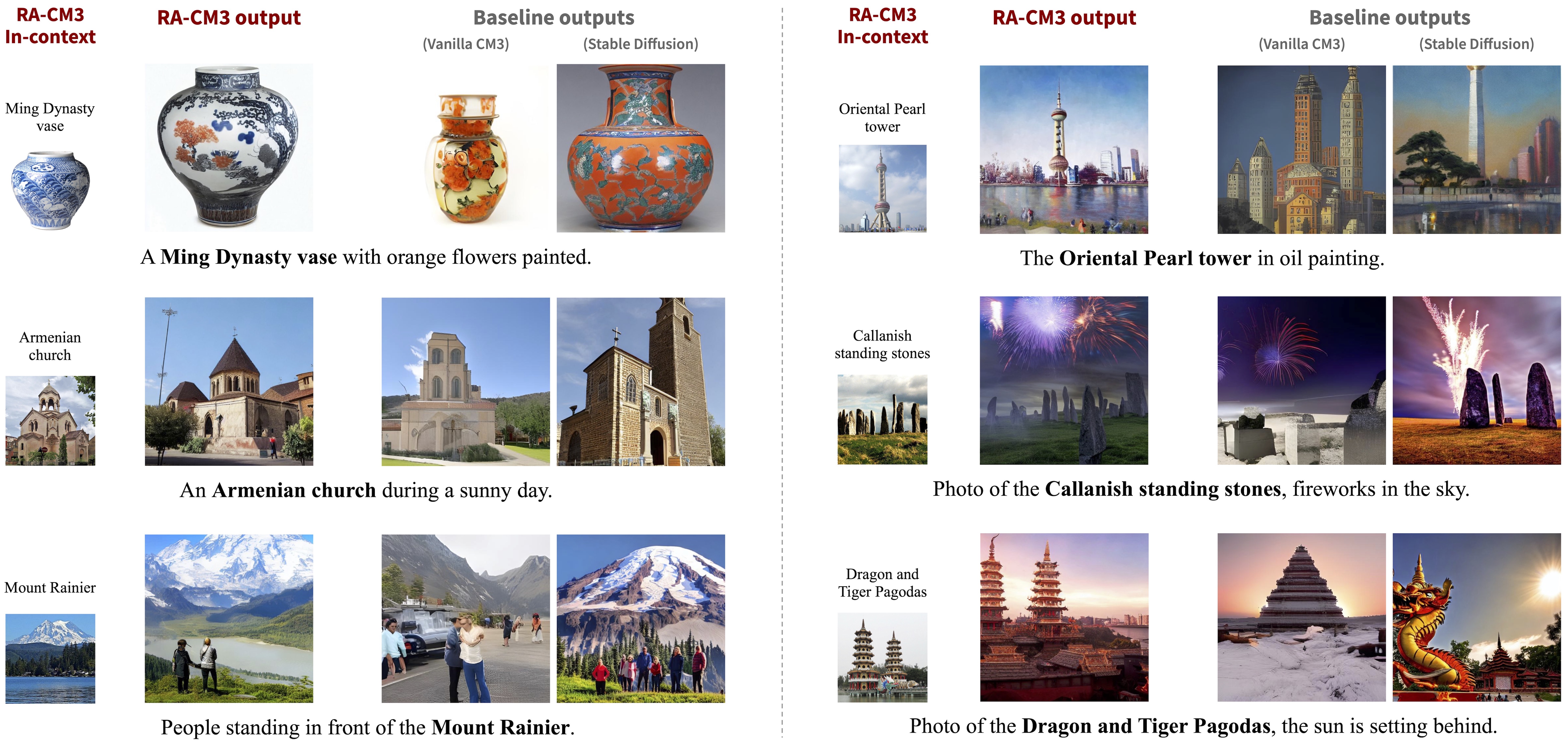}\vspace{-1mm}
    \caption{\textbf{Text-to-image generation involving world knowledge.}
    Our retrieval-augmented model (RA-CM3) can generate correct images from entity-rich captions thanks to the access to retrieved images in the context. For example, \methodname's outputs faithfully capture the visual characteristics of various entities (e.g., the shape and painting of Ming Dynasty vase, the amount of Callanish standing stones).
    On the other hand, baseline models without retrieval capabilities (vanilla CM3, Stable Diffusion) tend to struggle, especially when the caption involves rare entities (e.g., ``Ming Dynasty vase'', ``Oriental Pearl tower'', ``Dragon and Tiger Pagodas'').
    }
    \label{fig:knowledge_intensive}
\end{figure*}

\subsection{Main results}
\label{sec:exp-main_result}
\paragraph{Caption-to-image generation.}

Table \ref{tbl:coco_image_results} shows the caption-to-image generation performance on MS-COCO. The metric is FID score, where lower is the better.
Our retrieval-augmented CM3 achieves an FID score of 16 without finetuning, significantly outperforming the baseline CM3 with no retrieval (FID 29) and other models such as DALL-E (FID 28), which is 3x bigger than our model. This suggests that retrieval augmentation provides significant help in generating higher-quality images.

To also factor in training efficiency, 
Figure \ref{fig:compute} visualizes the image generation performance ($y$-axis: FID score) vs the amount of compute used in model training ($x$-axis: normalized A100 GPU hours) for our \methodname model and baseline models. We find that existing models in the autoregressive Transformer paradigm follow a negatively sloped line in this chart (the blue dots and line in Figure \ref{fig:compute}). \methodname is located significantly below this line, i.e., obtaining a better FID with less training compute. This suggests that the proposed retrieval-augmented method achieves significantly better training efficiency than existing works.

Our intuition is that retrieval augmentation allows the model to focus on learning how to use the retrieved documents in the context rather than fitting all the documents into the parameters of the model, speeding up the training process.

\paragraph{Image-to-caption generation.}

Table \ref{tbl:coco_caption_results} shows the image-to-caption generation performance on MS-COCO, with no finetuning.
The metric is the CIDEr score, where the higher is the better.
Our retrieval-augmented CM3 achieves a CIDEr score of 89, significantly outperforming the baseline CM3 with no retrieval (CIDEr 72). Moreover, \methodname outperforms other strong models such as Parti (20B parameters) and Flamingo (3B; 4-shot), despite using just $\sim$3B parameters and 2-shot in-context examples.

These results confirm that our model can perform both image and text generation well, offering the first unified retrieval-augmented multimodal model (Table \ref{tbl:method_comparison}).

\subsection{Analysis}
We analyze the scaling laws of \methodname in \S \ref{sec:exp-scaling_law} and the key design choices of \methodname in \S \ref{sec:exp-method-design}.

\begin{figure}[!t]
    \hspace{-3mm}
    \includegraphics[width=0.51\textwidth]{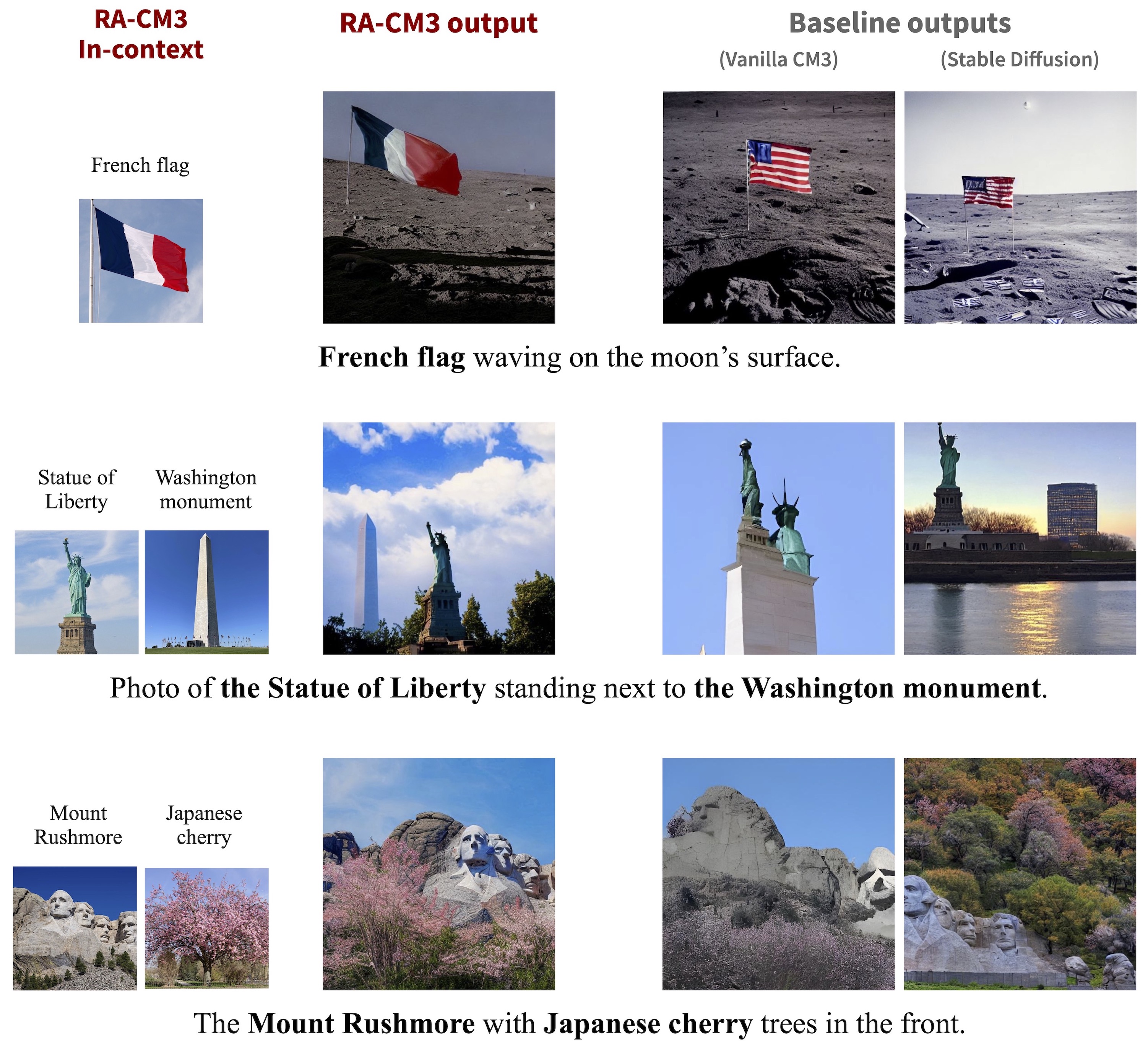}\vspace{-1mm}
    \caption{\textbf{Text-to-image generation involving rare \textit{composition} of knowledge.}
    Our retrieval-augmented model (\methodname) can generate faithful images from captions that contain a rare or unseen composition of entities (e.g., ``French flag'' + ``moon'', ``Mount Rushmore'' + ``Japanese cherry'').
    On the other hand, baseline models without retrieval capabilities (vanilla CM3, Stable Diffusion) tend to struggle on these examples, e.g., generate a US flag instead of a French flag on the moon.
    }\vspace{-1mm}
    \label{fig:rare_composition}
\end{figure}

\begin{figure}[!t]
    \centering
    \includegraphics[width=0.44\textwidth]{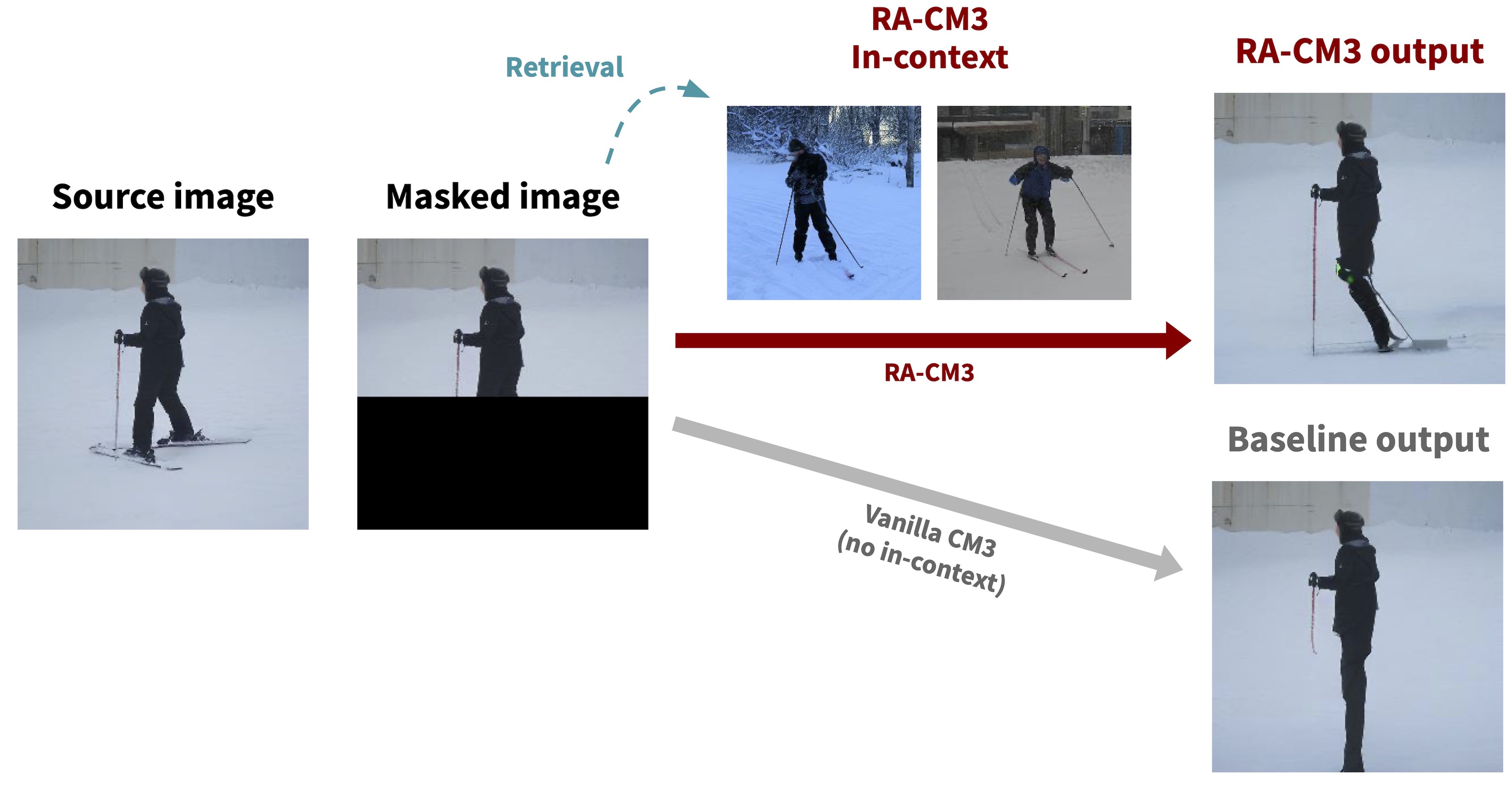}
    \vspace{-2mm}
    \caption{\textbf{Our model can perform better image infilling.}
    Infilling an image requires world knowledge, e.g., to recover the masked patches of the above image, the model needs to know about skiing. While the vanilla CM3 (no retrieval) tends to simply infill legs, our \methodname (with retrieval) successfully recovers both legs and skis. 
    }\vspace{-1mm}
    \label{fig:image_infill}
\end{figure}

\begin{figure}[!t]
    \centering
    \includegraphics[width=0.44\textwidth]{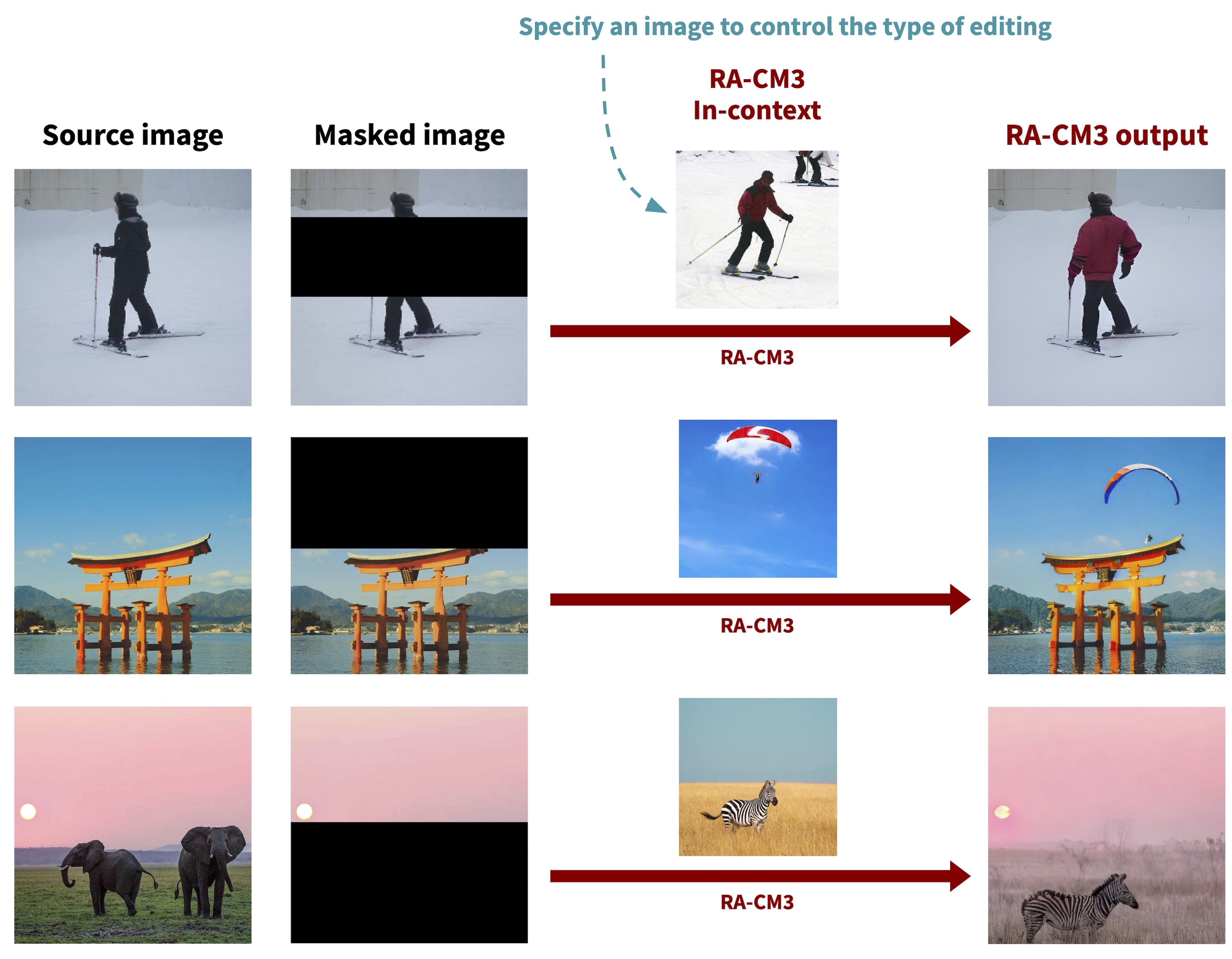}
    \vspace{-2mm}
    \caption{\textbf{Our model can perform image editing.} 
    Instead of using retrieved examples in our \methodname's context (Figure \ref{fig:image_infill}), we can also intervene and manually specify the in-context examples to control image infilling. For instance, we can place an image of a person wearing a red jacket in the context to edit the black jacket in the original image to be red (Figure top). 
    }
    \vspace{1mm}
    \label{fig:image_edit}
\end{figure}

\begin{figure}[!t]
    \hspace{-3mm}
    \includegraphics[width=0.51\textwidth]{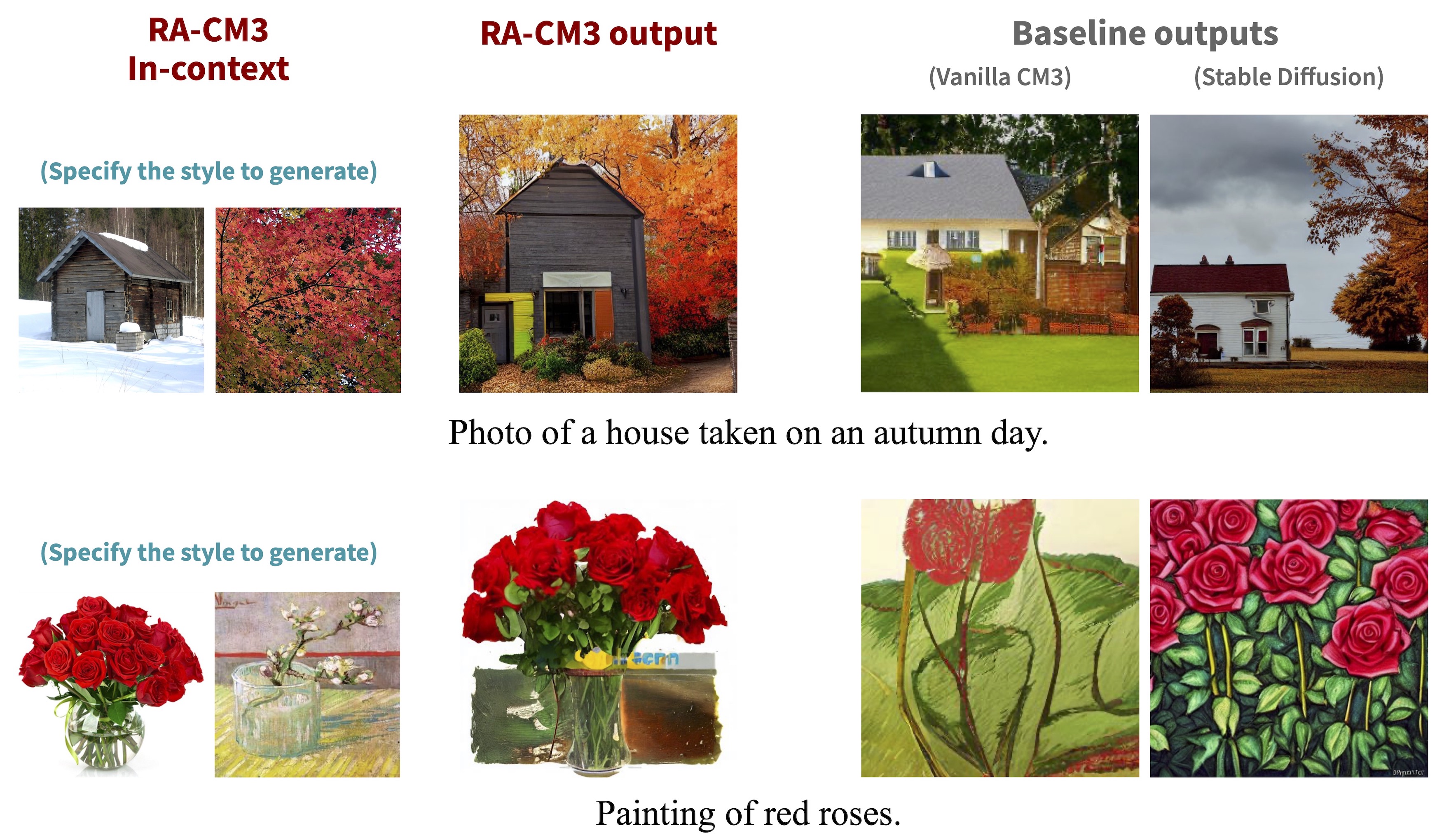}\vspace{-3mm}
    \caption{\textbf{Controllable image generation.}
    Our \methodname model can control the style of caption-to-image generation by prepending demonstration examples in the generator's context. For instance, when generating an image of ``a house taken on an autumn day'' (Figure top), we can specify a concrete style by
    providing demonstration images (e.g., image of a triangular wooden house and image of orange autumn leaves background). Consequently, \methodname generates an image that follows the visual characteristics of these in-context images.
    }
    \label{fig:control_gen}
\end{figure}

\begin{figure}[!t]
    \centering
    \hspace{-2mm}\includegraphics[width=0.495\textwidth]{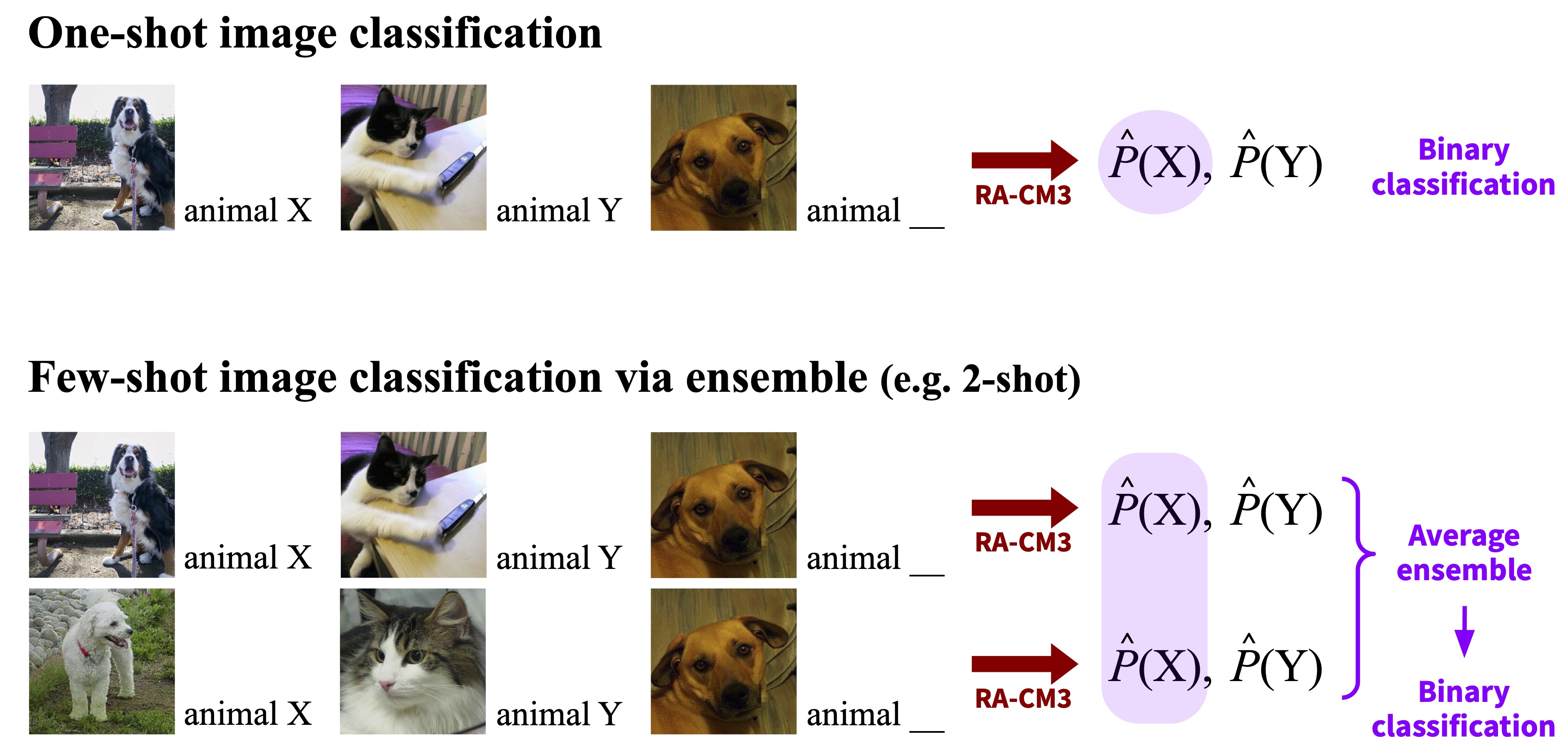}
    \vspace{-0mm}
    \begin{center}
    \setlength\tabcolsep{6pt}
    \resizebox{0.32\textwidth}{!}{
    \begin{tabular}{lcccc}
    \toprule
    \multicolumn{1}{l}{\multirow{2}{*}{\textbf{Model}}} & \multicolumn{4}{c}{\textbf{$\boldsymbol{k}$-shot Accuracy}}\\
    \cmidrule{2-5}
    & $k\!=\!1$ & $2$ & $4$ & $8$ \\
    \midrule
    Baseline CM3 &  0.53 & 0.50 & 0.56 & 0.56\\
    \textbf{\methodname (Ours)}~~~~~  &  \textbf{0.78} & \textbf{0.79} & \textbf{0.86} & \textbf{0.9}\\
    \bottomrule
    \end{tabular}
    }
    \end{center}
    \vspace{0mm}
    \caption{\textbf{Our model performs one/few-shot image classification via in-context learning}. 
    To assess the in-context learning ability, we consider a binary image classification task with non-semantic labels (e.g., ``animal X'' and ``animal Y'' instead of ``dog'' and ``cat''). 
    For one-shot classification (Figure top), we feed into the model one pair of demonstration examples, followed by a test example (\texttt{[test\,\,image]}, ``animal\,\_''), for which we predict the probability of ``X'' and ``Y''. 
    For $k$-shot classification (Figure middle), we repeat the above procedure $k$ times, each using a different pair of demonstration examples, and take the average ensemble of the predicted probability (``X'' and ``Y'') across the $k$ passes.\\[1mm]
    The table (Figure bottom) shows the results of $k$-shot classification accuracy, with $k=1,2,4,8$. Across all $k$'s, our \methodname improves on the baseline CM3 by large margins. Increasing $k$ consistently improves accuracy for the $k$ values above.
    }
    \label{fig:image_one_shot}
\end{figure}

\section{Qualitative results}
\label{sec:qualitative-results}
We show novel qualitative capabilities of our \methodname, such as knowledge-intensive multimodal generation (\S \ref{sec:result-knowledge}) and multimodal in-context learning (\S \ref{sec:result-image_infill}, \ref{sec:result-controlled}, \ref{sec:result-oneshot}). 
While GPT-3 \cite{brown2020language} and Flamingo \cite{alayrac2022flamingo} showed in-context learning for text-to-text or image-to-text generation, we show that \methodname can do in-context learning for both text (\S \ref{sec:result-oneshot}) and image (\S \ref{sec:result-image_infill}, \ref{sec:result-controlled}) generation.

\subsection{Knowledge-intensive multimodal generation}
\label{sec:result-knowledge}
Because of the retrieval capability, \methodname is especially good at tasks that require world knowledge or composition of knowledge (knowledge-intensive generation).
Figure \ref{fig:knowledge_intensive}, \ref{fig:rare_composition} show example outputs from \methodname. For each caption, the output images were obtained by sampling 256 images from the model and then re-ranking them using the CLIP score with respect to the input caption. We then apply an off-the-shelf super-resolution tool \cite{rombach2022high}.

\textbf{World knowledge.} 
Figure \ref{fig:knowledge_intensive} shows model outputs for caption-to-image generation that involves world knowledge (e.g., specific entities).
We find that our \methodname model can generate correct images from entity-rich captions thanks to the access to retrieved images in the context. For example, \methodname's outputs faithfully capture the visual characteristics of various entities (e.g., the shape and painting of Ming Dynasty vase, the amount of Callanish standing stones).
On the other hand, baseline models without retrieval capabilities (vanilla CM3, Stable Diffusion) tend to struggle, especially when the caption involves rare entities (e.g., ``Ming Dynasty vase'', ``Oriental Pearl tower'', ``Dragon and Tiger Pagodas'').

\textbf{Composition of knowledge.} 
Figure \ref{fig:rare_composition} shows model outputs for caption-to-image generation that involves rare \textit{composition} of knowledge. 
We find that our retrieval-augmented model can generate faithful images from captions that contain a rare or unseen composition of entities (e.g., ``French flag'' + ``moon'', ``Mount Rushmore'' + ``Japanese cherry'').
On the other hand, baseline models without retrieval capabilities (vanilla CM3, Stable Diffusion) tend to struggle on these examples, e.g., generate a US flag instead of a French flag on the moon (Figure \ref{fig:rare_composition} top). This is likely because the US flag was the most common flag that co-occurred with the moon in the training data.

\subsection{Image infilling and editing}
\label{sec:result-image_infill}

Because our model builds on CM3, it can also perform \textit{infilling}.\footnote{To perform image infilling, the model is given ``\texttt{[unmasked part of image] <mask> [unmasked part of image]}'' as the prompt and generates the \texttt{<mask>} part as the completion. The final output image is constructed by plugging the generated output to the \texttt{<mask>} part of the input.} Figure \ref{fig:image_infill} shows that our \methodname can perform improved image infilling because of the retrieval capability. Infilling an image requires world knowledge, e.g., to recover the masked patches of the image in Figure \ref{fig:image_infill}, the model needs to know about skiing. While the vanilla CM3 (no retrieval) tends to simply infill legs, \methodname (with retrieval) successfully recovers both legs and skis. 

Moreover, instead of using retrieved examples in the \methodname context, we can also intervene and manually specify the in-context examples to control image infilling. Figure \ref{fig:image_edit} shows examples. For instance, we can place an image of a person wearing a red jacket in the context to edit the black jacket in the original image to be red (Figure \ref{fig:image_edit} top).

\subsection{Controlled image generation}
\label{sec:result-controlled}

Controlled generation---controlling the behavior of models in generation (e.g., style of outputs)---is a key problem in generative models \cite{keskar2019ctrl, li2019controllable}.

Our \methodname can control the style of caption-to-image generation by prepending demonstration examples in the generator's context (Figure \ref{fig:control_gen}).
For instance, when generating an image for ``Photo of a house taken on an autumn day'' (Figure \ref{fig:control_gen} top), we can specify a concrete style by providing demonstration images (e.g., an image of a triangular wooden house and an image of orange autumn leaves background). Consequently, \methodname can generate an image that actually follows the visual characteristics of these in-context images. 
This is a very useful capability because we can control generation not only via text (captions) but also via image demonstrations---especially helpful when some visual characteristics we want to specify might be difficult to express in text.

Moreover, the finding that \methodname can use in-context examples for controlled generation suggests that it has acquired a form of \textbf{multimodal in-context learning} ability. Our intuition is that because the \methodname generator has seen relevant multimodal documents prepended to the main document in context during retrieval-augmented training, it has learned how to use in-context examples effectively.

\subsection{One-shot and few-shot image classification}
\label{sec:result-oneshot}
So far we have seen \methodname's in-context learning behavior for image generation (\S \ref{sec:result-controlled}). Here we study its in-context learning ability for image-to-text generation, through one-shot and few-shot image classification.

Figure \ref{fig:image_one_shot} illustrates the experiment.
To assess the true in-context learning ability that factors out prior knowledge of the model, we consider a binary image classification task with non-semantic labels (e.g., ``animal X'' and ``animal Y'' instead of ``dog'' and ``cat''). Specifically, we use ImageNet \cite{deng2009imagenet} to construct such evaluation sets where each class (e.g., animal X or Y) contains the same number of test images (e.g., 100 images).
For one-shot classification (Figure \ref{fig:image_one_shot} top), we feed into the model one pair of demonstration examples (\texttt{[image\,X]}, ``animal X'', \texttt{[image\,Y]}, ``animal Y''), followed by a test example (\texttt{[test\,\,image]}, ``animal\,\_''), for which we predict the probability of ``X'' and ``Y''. 
For $k$-shot classification (Figure \ref{fig:image_one_shot} middle), we repeat the above procedure $k$ times, each using a different pair of demonstration examples, and take the average ensemble of the predicted probability (``X'' and ``Y'') across the $k$ passes.\footnote{An alternative way to use $k$-shot examples could be to prepend all the $k$ pairs of demonstrations directly in \methodname's context, but this would take a significant sequence length in Transformer and might not be easy to scale. 
We find that the ensemble-based method performs well empirically, and comes with benefits such as being more scalable (parallel runs of shorter-length passes) and principled (agnostic to the order of the $k$ examples).}

The table in Figure \ref{fig:image_one_shot} bottom shows the results of $k$-shot (binary) classification accuracy, with $k=1,2,4,8$.
Across all $k$'s, our \methodname obtains significantly improved accuracy over the baseline CM3, which were not trained with retrieved documents in context. In particular, \methodname already performs reasonably well at one-shot (0.78 accuracy at $k=1$). This result suggests that \methodname has acquired a strong in-context learning ability, especially given that we use non-semantic labels for image classes in this evaluation.
Moreover, we find that increasing $k$ consistently improves accuracy for the $k$ values above (0.90 accuracy at $k=8$). 
This observation suggests that ensemble is an effective method to increase the number of in-context examples to provide for the model.

\section{Conclusion}
We presented a retrieval-augmented multimodal model that can retrieve and refer to an external memory for generating images and text. Specifically, we implemented a multimodal retriever using the pretrained CLIP and designed a retrieval-augmented generator using the CM3 architecture. 
Our resulting model, named \methodname, outperforms existing multimodal models on both image and caption generation tasks, while requiring much less training compute.
Moreover, \methodname exhibits novel capabilities such as knowledge-intensive image generation and multimodal in-context learning. 

This work aims to offer a general and modular retrieval augmentation framework for multimodal models. We believe this opens up various exciting research avenues, such as improving the multimodal retriever and generator, extending modalities beyond image and text, and further investigating multimodal prompting and in-context learning.

\section*{Acknowledgements}
We thank members of the Meta AI team, Stanford P-Lambda and SNAP groups, as well as our anonymous reviewers for providing valuable feedback.

\bibliography{main}
\bibliographystyle{icml2023}

\newpage
\appendix
\section{Ethics and societal impact}
\label{sec:ethics}
Multimodal generative models, including our \methodname and other models like DALL-E, Parti and Stable Diffusion, are typically trained on large, noisy image-text datasets collected from the web. These datasets may contain biases about race, gender and other demographic attributes, and unsafe content such as pornography and violence \cite{birhane2021multimodal}. 
We performed extensive data filtering to remove problematic content in training data following existing works (\S \ref{sec:exp-train_setup}), but it may still be possible that \methodname outputs problematic text or images. 
\methodname is a \textbf{research prototype}, and we do not encourage using it in high-risk or sensitive domains or for generating images of people.
For a more general, detailed discussion on the ethical considerations of multimodal generative models, we refer readers to e.g., \citet{denton2021ethical}.

We also highlight the potential societal benefits of our retrieval-augmented multimodal model. First, as our model requires much less compute for training than existing models (\S \ref{sec:exp-main_result}), it can provide \textbf{energy savings}.
Second, retrieval helps capture long-tail knowledge (e.g., rare entities or minority groups), which can contribute to more \textbf{fair} multimodal models.
Third, retrieval naturally provides the provenance of knowledge, offering better {interpretability} and \textbf{explainability} about model predictions.
Retrieval augmentation also helps make image/text generation \textbf{faithful} to the retrieved evidence documents (\S \ref{sec:result-knowledge}), potentially helping reduce unintentionally fake or hallucinated outputs.

\section{Related work}
\label{sec:related_work}

\paragraph{Vision-language multimodal models.}
Various models have been developed for text-to-image generation. Typically, these models are autoregressive Transformer-based, e.g., DALL-E \cite{ramesh2021zero} and Parti \cite{yu2022scaling}, or diffusion-based, e.g., Imagen \cite{saharia2022photorealistic}, DALL-E 2 \cite{ramesh2022hierarchical} and Stable Diffusion \cite{rombach2022high}.
Meanwhile, several works also study image-to-text generation \cite{cho2020x,wang2021simvlm}.
In particular, Flamingo \cite{alayrac2022flamingo} is a Transformer-based image-to-text generation model, with in-context learning ability.
Recently, CM3 \cite{aghajanyan2022cm3} provides a unified model that uses a Transformer to perform both text and image generation. To make use of this generality, we will build on CM3 to design our model.

While the above models have achieved strong performance on image and text generation, they store all their knowledge inside the model, which tends to require a lot of parameters (e.g., 10B) and training data (e.g., 1B images).
To address this limitation, we augment them with an ability to refer to relevant examples from an external memory when generating images/text.
With this augmentation, our model outperforms the existing models by using less training data (150M images), compute and parameters ($< \!30\%$) (\S \ref{sec:experiments}).

\paragraph{Retrieval-augmented language models.}
Retrieval augmentation has shown promise in NLP \cite{lewis2020retrieval}. To incorporate knowledge into a language model (LM), this line of work retrieves documents relevant to input text from an external memory, and lets the LM (generator) use the retrieved documents to make more informed predictions.
The external memory used is typically a collection of text passages \cite{hashimoto2018retrieve, khandelwal2019generalization, karpukhin2020dense, guu2020realm, lewis2020retrieval, lewis2020pre, yasunaga2022linkbert, borgeaud2022improving, shi2023replug}
or a structured knowledge base \cite{zhang2019ernie,Agarwal2021KnowledgeGB, xie2022unifiedskg,yasunaga2021qa,yasunaga2022dragon}.
Here we generalize the scope of the retrieval-augmented LM framework and consider \textit{multimodal} documents for both our input data and external memory, which can be arbitrary sequences of text and images.

\paragraph{Retrieval in multimodal models.}
Besides the retrieval augmentation for language models, recent works also study retrieval for computer vision models \cite{ashual2022knn,blattmann2022retrieval, gur2021cross,sarto2022retrieval,li2022comprehending,ramos2022smallcap, wang2022vqa}.
More recently, some concurrent works study retrieval in multimodal models:
Re-Imagen \cite{reimagen} performs diffusion-based caption-to-image generation using retrieved images;
MuRAG \cite{chen2022murag} performs natural language question answering using retrieved images.
While these works focus on generating a single modality, either text or image, we develop a general and unified model that can retrieve, encode, and generate \textit{both} images and text (Table \ref{tbl:method_comparison}).
Moreover, our retrieval-augmented training allows the generator model to acquire novel in-context learning ability such as controlled image generation (\S \ref{sec:qualitative-results}).

\section{Additional results}

\subsection{Intrinsic evaluation of CLIP-based retriever}
\label{sec:retrieval_eval}

\begin{table}[H]
\begin{center}
\scriptsize
\setlength\tabcolsep{4pt}
\resizebox{0.4\textwidth}{!}{
\begin{tabular}{lccc@{}}
\toprule
\multicolumn{1}{l}{\multirow{2}{*}{\textbf{Method}}} & \multicolumn{3}{c}{\textbf{Recall ($\uparrow$)}}\\
\cmidrule{2-4}
& @1 & @3 & @5 \\
\midrule
CLIP text-to-image retrieval & 48 & 65 & 78 \\
CLIP text-to-mixture retrieval & 61 & 77 & 85 \\
\midrule
CLIP image-to-text retrieval & 56 & 75 & 84 \\
CLIP image-to-mixture retrieval~~ & 78 & 84 & 87\\
\bottomrule
\end{tabular}
}\vspace{-1mm}
\caption{\textbf{Multimodal retrieval performance on MS-COCO}. We use the frozen pretrained CLIP. ``text-to-image retrieval'' and ``image-to-text retrieval'' use the CLIP text/image encoder as it is. ``text-to-mixture retrieval'' and ``image-to-mixture retrieval'' use our mixed-modal encoder based on CLIP (\S \ref{sec:approach-retrieval}). In all these cases, the CLIP-based retrieval method performs reasonably well.
}
\label{tbl:retrieval_eval}
\end{center}
\end{table}

\subsection{Scaling laws of \methodname}
\label{sec:exp-scaling_law}
\begin{figure}[!t]
    \hspace{-4mm}
    \includegraphics[width=0.52\textwidth]{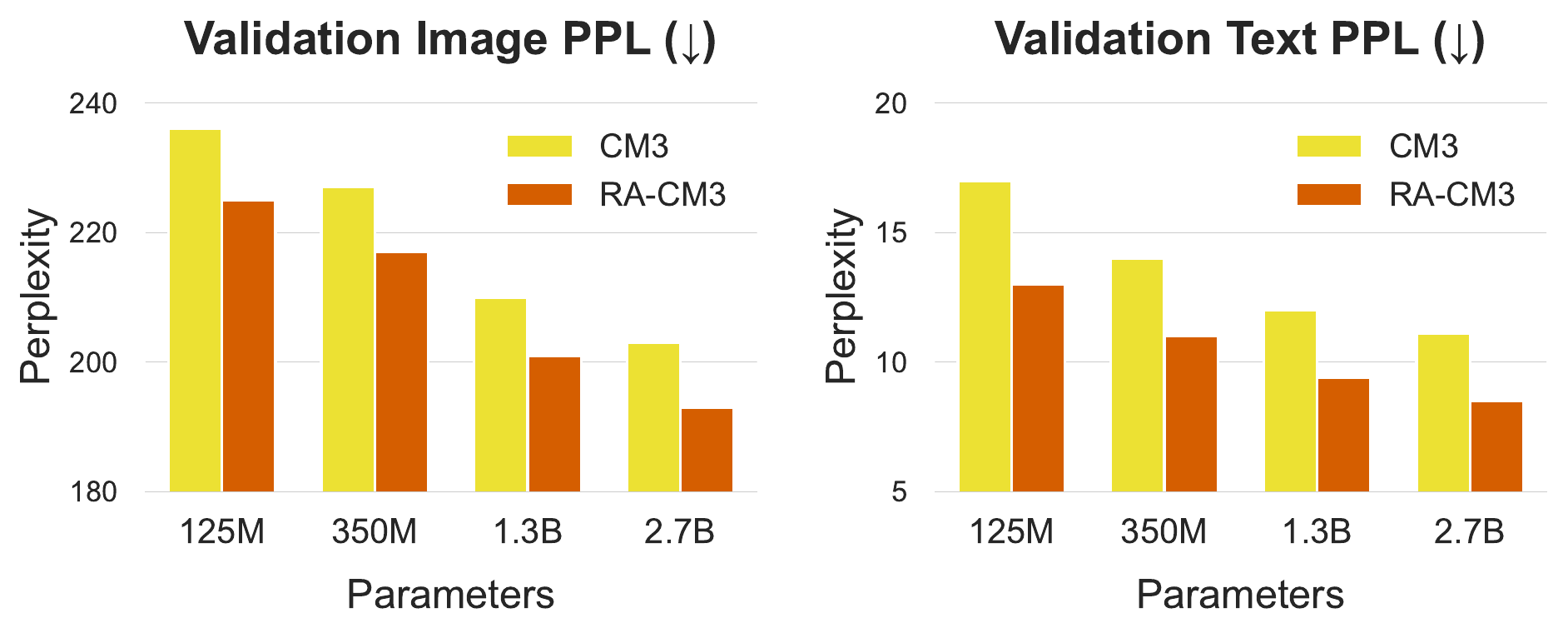}
    \vspace{-6mm}
    \caption{\textbf{Perplexity-based scaling laws for our \methodname model.} 
    We train \methodname and vanilla CM3 of various parameter counts using the same amount of compute, and evaluate perplexity on the held-out validation set of MS-COCO.
    \methodname provides consistent improvements over vanilla CM3 across different scales.
    }
    \label{fig:scaling}
\end{figure}

To study the scaling laws of retrieval augmentation for multimodal models, we train the retrieval-augmented CM3 and vanilla CM3 of various sizes (125M, 350M, 1.3B, 2.7B parameters) using the same amount of compute (two days on 256 GPUs), and then evaluate the models' perplexity on the MS-COCO validation set (Figure \ref{fig:scaling}).
We observe that \methodname provides consistent improvements over vanilla CM3 across different scales. We do not observe any diminishing returns in the 125M--2.7B range that we studied. This suggests that retrieval augmentation is also promising at a larger scale.

\begin{table}[!t]
\aboverulesep = 0.705mm \belowrulesep = 1.084mm
    \centering
    \scalebox{0.75}{
    \small
    \begin{tabular}{rlcc}
     \toprule
    \textbf{Method design}\!\!\vrule width 0pt depth 3pt height 8pt & \textbf{Choice} & \!\!\textbf{Image ppl ($\downarrow$)}\!\! & \!\!\textbf{Text ppl ($\downarrow$)}\!\! \\
    \midrule
    \!\!\multirow{4}{*}{
    \begin{tabular}{@{}l@{}}\vrule width 0pt depth 0pt height 8pt \hspace{0em}Retrieval relevance\\[-0.1em]
    \hspace{2em}(\S \ref{sec:approach-retrieval-strategy})\end{tabular}
    }\!\! 
    & Random at train \& infer &  246  & 23 \\[0.0em]
    & Retrieve at train, random at infer\!\! &  246  & 24 \\[0.0em]
    & Random at train, retrieve at infer\!\! &  243  & 18 \\[0.0em]
    & Retrieve at train \& infer (\textbf{final})\!\!  & \textbf{227} & \textbf{13}  \\
    \midrule
    \!\!\multirow{2}{*}{
    \begin{tabular}{@{}l@{}}\vrule width 0pt depth 0pt height 8pt \hspace{0em}Retrieval modality\\[-0.1em]
    \hspace{2em}(\S \ref{sec:approach-retrieval-strategy})\end{tabular}
    }\!\! & Only image or only text & 234 & 15  \\[0.1em]
    & Multimodal document (\textbf{final})\!\!  & \textbf{227} & \textbf{13}  \\
    \midrule
    \!\multirow{4}{*}{
    \begin{tabular}{@{}l@{}}\vrule width 0pt depth 0pt height 8pt \hspace{0em}Retrieval diversity\\[-0.1em]
    \hspace{2em}(\S \ref{sec:approach-retrieval-strategy})\end{tabular}
    }\!\! & Simply take top $K$ & 244 & 17 \\[0.1em]
    &  Avoid redundancy & 235 & 15 \\
    &  
    \begin{tabular}{@{}l@{}}\vrule width 0pt depth 0pt height 9pt \hspace{0em}Avoid redundancy\\[-0.2em] 
    \hspace{0.56em}+ Query dropout\end{tabular} (\textbf{final})\!\!\!\!  &
    \textbf{227} & \textbf{13}  \\
    \midrule
    \!\multirow{4}{*}{
    \begin{tabular}{@{}l@{}}\vrule width 0pt depth 0pt height 8pt \hspace{0em}Generator training\\[-0.1em]
    \hspace{1.2em}(Equation \ref{eq:loss})\end{tabular}
    }\!\! & $\alpha=0$ & 239  & 17 \\
     & $\alpha=1$ & 240 & 17 \\
     & $\alpha=0.3$ & 231 & 14 \\
     & $\alpha=0.1$ (\textbf{final}) & \textbf{227} & \textbf{13} \\
    \bottomrule
    \end{tabular}}
    \vspace{-0mm}
    \caption{
    \textbf{Analysis of our method's design choices.} As the metric, we use the perplexity of image/text generation on the MS-COCO validation set. We find that key methods to achieve the best performance are: ensure \textit{relevance} in retrieved documents (table top); retrieve \textit{multimodal} documents instead of only images or text (table second from top); encourage \textit{diversity} in retrieved documents during training (table second from bottom); and train the token prediction loss for \textit{both} the main input document and the retrieved documents, in particular, with a weight of $\alpha=0.1$ (table bottom).\\
    Note that images naturally have higher perplexity than text, as also observed in prior works (e.g., \citealt{aghajanyan2022cm3}).
    }
    \label{tab:ablation}
\end{table}
\subsection{Analysis of \methodname designs}
\label{sec:exp-method-design}
We analyze key design choices of the retrieval-augmented multimodal model, such as the strategies of retrieval (\S \ref{sec:approach-retrieval}) and generator training (\S \ref{sec:approach-generator}).
Here, our experiments used a 2.7B parameter-model, trained for a day on 256 GPUs.

\textbf{Retrieval relevance} (Table \ref{tab:ablation} top).
A main contribution of our work is retrieval-augmented training of the generator (\S \ref{sec:approach-generator}).
While our final \methodname prepends documents retrieved by our CLIP-based retriever at both train and inference time (``Retrieve at train \& infer'' row in the table), 
one natural baseline is to train the model using random documents without retrieval (i.e., vanilla CM3) but use retrieved documents at inference time (``Random at train, retrieve at infer'').
This baseline leads to a significant performance drop, suggesting that having relevant documents in context is crucial for model training. 
We also study other baselines, such as using retrieved documents at train time but random documents at inference time, or using random documents at both train and inference times. Both result in significant performance drops. These results confirm that relevance is a crucial factor in retrieval at both train and inference times.

\textbf{Retrieval modality} (Table \ref{tab:ablation} second from top). 
While existing works on retrieval \cite{reimagen} typically retrieve either an image or text only for the generator, our retriever based on a mixed-modal encoder (\S \ref{sec:approach-retrieval}), can retrieve a multimodal document that consists of both images and text. We find that retrieving multimodal documents performs better than retrieving only images or text. Our intuition is that a multimodal document can be more informative because the text and image within it can contextualize each other.

\textbf{Retrieval diversity} (Table \ref{tab:ablation} second from bottom). 
As discussed in \S \ref{sec:approach-retrieval-strategy}, encouraging diversity in retrieved documents is important.
Simply taking the top $K$ (e.g., 2) from the list of candidate documents sorted by the retriever scores leads to poor performance---in fact, slightly worse than the baseline with no retrieval augmentation. Our first technique which avoids redundancy in retrieved documents leads to significant performance improvement.
We also find that the second technique, Query Dropout, which encourages more diversity in retrieval during training leads to a further boost in evaluation performance.

\textbf{Generator training} (Table \ref{tab:ablation} bottom). 
A key design of our generator is that we optimize token prediction loss jointly for the main input document and the retrieved documents, with a weighting $\alpha$ (\S \ref{sec:approach-generator}; Equation \ref{eq:loss}). 
Existing retrieval-augmented models typically optimize loss for the main document only ($\alpha=0$), but we find that joint optimization ($\alpha>0$) facilitates training and improves performance. We find that $\alpha=0.1$ works well. 
Setting $\alpha$ to be too large (e.g., $\alpha=1$) hurts training because this would place too much weight on modeling retrieved documents instead of the main document.

\section{Additional discussions}

\subsection{Fair comparison of the retrieval-augmented model and non-retrieval-augmented model}

\paragraph{Experiments:}
Our baseline is the vanilla CM3 with no retrieval augmentation.
To make a fair comparison between the retrieval-augmented model (\methodname) and the baseline, \methodname is trained using the same generator architecture, same training data, and same amount of compute as the vanilla CM3 (\S \ref{sec:exp-train_setup}). \methodname's external memory used for retrieval also consists of the same training data. 
Thus, we ensured that no additional data or training compute is used for the retrieval-augmented model compared to the non-retrieval-augmented models.
Under this controlled experiment, \methodname substantially outperforms the vanilla CM3 in image and text generation (Table \ref{tbl:coco_image_results}, \ref{tbl:coco_caption_results}).
Figure \ref{fig:scaling} further indicates that \methodname with fewer parameters (1.3B) can also outperform the vanilla CM3 with more parameters (2.7B).
We also include the ``Retrieval Baseline'' in Table \ref{tbl:coco_image_results} and \ref{tbl:coco_caption_results}, which simply returns the retrieved images or text as model outputs. \methodname outperforms this retrieval baseline.

\paragraph{Usage:}
Both the retrieval-augmented model and non-retrieval-augmented models take the same input from users, e.g., a source caption for image generation or a source image for caption generation. In the retrieval-augmented model, the retriever will automatically take this input prompt, fetch relevant images/text, and add them to the context of the generator, so no additional input is needed from the user. 
Of course, the user may also intervene and self-specify in-context examples for the generator, so the retrieval-augmented model provides more flexibility and controllability for users (\S \ref{sec:result-controlled}). 
The retrieval step can be performed efficiently using FAISS (\S \ref{sec:exp-train_setup}) in less than a second.

Thus, while the retrieval-augmented model operates within the same (fair) task definition as non-retrieval-augmented models, it opens up various benefits and new capabilities, such as improved explainability, faithfulness, and controllability in generation (\S \ref{sec:ethics}).

\subsection{Taking an existing model (e.g.~vanilla CM3) and finetune it with retrieval-augmentation, instead of training the retrieval-augmented model (\methodname) from scratch}
In practice, finetuning from the vanilla model can save compute for training \methodname and is useful. The reason we trained \methodname and vanilla CM3 both from scratch in our main experiments was to make a fair comparison between them by training with the same amount of compute, and to systematically study the effect of retrieval augmentation.

\subsection{How the number of retrieved documents used for the generator ($\boldsymbol{K}$) was set}

We set $K$ to be up to $2$ (\S \ref{sec:exp-train_setup}), primarily in consideration of the Transformer sequence length. Using the recent image tokenizer (e.g., \citealt{esser2021taming}), each image is mapped to 1K tokens. Hence, the concatenation of ${K\!=\!2}$ retrieved documents and the main input document takes 3--4K tokens in total in Transformer. Increasing the Transformer sequence length beyond 4K incurs a significant burden in computation and GPU memory, so we decided on ${K\!=\!2}$. ${K\!=\!2}$ also worked reasonably well in practice (\S \ref{sec:exp-main_result}).

During inference, we may do ensemble (see \S \ref{sec:result-oneshot}) and take more than two retrieved documents in total into account. We conducted an analysis of varied $K\!=\!1,2,4,8$ in the MS-COCO caption-to-image generation evaluation (Table \ref{tbl:PPL_vs_K}). We find that $K\!=\!2$ worked the best in this experiment. Our intuition is that most of the MS-COCO captions involve 1--2 objects, so retrieving the top two multimodal documents may be sufficient for generating corresponding images. It is an interesting future research to investigate larger $K$'s on image generation tasks that involve more entities/objects.

\begin{table}[!h]
    \begin{center}
    \setlength\tabcolsep{6pt}
    \scalebox{0.8}{
    \begin{tabular}{lcccc}
    \toprule
    \multicolumn{1}{l}{\multirow{2}{*}{\textbf{Model}}} & \multicolumn{4}{c}{\textbf{Image Perplexity ($\downarrow$)}}\\
    \cmidrule{2-5}
    & $K\!=\!1$ & $2$ & $4$ & $8$ \\
    \midrule
    \textbf{\methodname}~~~~~ & 228 & 227 & 228 & 232 \\
    \bottomrule
    \end{tabular}
    }
    \end{center}
    \vspace{-3mm}
    \caption{\textbf{MS-COCO caption-to-image generation performance} when the number of retrieved multimodal documents ($K$) is varied.
    }
    \label{tbl:PPL_vs_K}
\end{table}



\end{document}